\documentclass[10pt,twocolumn,letterpaper]{article}

 \usepackage{cvpr}              
\usepackage{graphicx}
\usepackage{amsmath}
\usepackage{amssymb}
\usepackage{booktabs}
\usepackage{multicol, multirow}

\renewcommand{\thefootnote}{\fnsymbol{footnote}}
\newcommand\blfootnote[1]{%
  \begingroup
  \renewcommand\thefootnote{}\footnote{#1}%
  \addtocounter{footnote}{-1}%
  \endgroup
}

\usepackage[dvipsnames]{xcolor}

\definecolor{cvprblue}{rgb}{0.21,0.49,0.74}
\usepackage[pagebackref,breaklinks,colorlinks,citecolor=cvprblue]{hyperref}

\def\paperID{1558} 
\def\confName{CVPR}
\def\confYear{2024}

\title{Mesh Represented Recycle Learning for 3D Hand Pose and Mesh Estimation}

\author{Bosang Kim$^\dagger$, Jonghyun Kim$^\dagger$, Hyotae Lee, Lanying Jin, Jeongwon Ha, \\
    Dowoo Kwon, Jungpyo Kim, Wonhyeok Im, KyungMin Jin, and Jungho Lee \\
LG Electronics\\
19, Yangjae-daero 11gil, Seocho-Gu, Seoul, Republic of Korea, 06772\\
\tt\small \{bosang1.kim, jonghyun0.kim, hyotae.lee, lanying.jin, jeongwon.ha, \\
   \tt\small dowoo.kwon,  jungpyo.kim, wonhyeok.im, kyungmin.jin, jungo.lee\}@lge.com}

\begin{document}
\maketitle
\begin{abstract}
   In general, hand pose estimation aims to improve the robustness of model performance in the real-world scenes. However, it is difficult to enhance the robustness since existing datasets are obtained in restricted environments to annotate 3D information. Although neural networks quantitatively achieve a high estimation accuracy, unsatisfied results can be observed in visual quality. This discrepancy between quantitative results and their visual qualities remains an open issue in the hand pose representation. To this end, we propose a mesh represented recycle learning strategy for 3D hand pose and mesh estimation which reinforces synthesized hand mesh representation in a training phase. To be specific, a hand pose and mesh estimation model first predicts parametric 3D hand annotations (i.e., 3D keypoint positions and vertices for hand mesh) with real-world hand images in the training phase. Second, synthetic hand images are generated with self-estimated hand mesh representations. After that, the synthetic hand images are fed into the same model again. Thus, the proposed learning strategy simultaneously improves quantitative results and visual qualities by reinforcing synthetic mesh representation. To encourage consistency between original model output and its recycled one, we propose self-correlation loss which maximizes the accuracy and reliability of our learning strategy.  Consequently, the model effectively conducts self-refinement on hand pose estimation by learning mesh representation from its own output. To demonstrate the effectiveness of our learning strategy, we provide extensive experiments on FreiHAND dataset. Notably, our learning strategy improves the performance on hand pose and mesh estimation without any extra computational burden during the inference. 
   \blfootnote{$\dagger$ Both authors are equally contributed.}
\end{abstract}    
\section{Introduction}
\label{sec:intro}
The goal of hand pose and shape estimation is to predict precise 3D keypoint coordinates and mesh vertices. This task is crucial in vision applications such as virtual reality, robotics, and human-computer interaction. Over the years, many researchers have explored various approaches to address this challenging task. 

Hand pose estimation (HPE) can be categorized into two approaches: non-parametric and parametric model-based methods. The non-parametric methods directly estimate hand keypoints from input images without post-processing steps \cite{wan2018dense, hand2019survey, Iqbal_2018_ECCV}. However, these methods occasionally yield anatomically unpredictable hand structures due to the lack of structured representation for hand poses \cite{gomez2019large}. Moreover, these methods are vulnerable to occluded and ambiguous hand poses in a scene \cite{ye2018occlusion}. The reason is that non-parametric methods do not incorporate explicit modeling of the geometric hand structure \cite{Baek_2019_CVPR}. Thus, it is difficult to recover missing or ambiguous visual information from aforementioned hand poses. This limitation causes inaccurate hand keypoints prediction and reduces robustness in real-world. In contrast, parametric model-based methods represent a hand pose with a set of deformable parameters which contains structural information of hand poses \cite{mano}. Specifically, hand parameters describe a hand shape with 3D vertices and articulation movements, that ensure physically feasible poses. Although the parametric model-based methods are robust to learn geometric information of hand poses, these methods require elaborate annotated data of hand poses and shapes to learn their parameters \cite{ho3d}. Nevertheless, it is laborious and time-consuming to acquire large and diverse hand data with accurate annotations. This data scarcity causes a significant obstacle for training robust parametric model-based HPE methods \cite{ohkawa2023efficient}. Consequently, it is difficult to generalize these methods in unseen poses and various backgrounds \cite{liu2021semi}. It means that unsatisfied results are observed in visual quality of hand meshes despite achieving higher accuracy in HPE \cite{moon2020i2l, grady2021contactopt}. This discrepancy between quantitative results and their visual qualities remains as the main concern in this task.

To resolve these issues, we propose a novel learning approach, called recycle learning, that capitalizes on the synergy between real-world and synthetic data. In the proposed recycle learning strategy, a HPE network first estimates 3D hand annotations with real-world hand images in a training phase while establishing a foundation for accurate hand pose estimation. After that, to alleviate the discrepancy between quantitative performance and its visual result, we employ a subsequent training phase utilizing estimated hand annotations in the previous training phase. In this second training phase, synthetic hand images are generated from the estimated hand annotations, and provided to the same network again. Thus, this training procedure facilitates the HPE network to learn self-estimated mesh representations. Consequently, the recycle learning strategy mitigates the limitation of 3D annotated data, and enhances the robustness of hand pose and mesh estimation in generalizability. In addition, to ensure consistency between the outputs of the first and second learning phases, we introduce a self-correlation loss. By minimizing this loss, the network is encouraged to produce coherent estimation results throughout two separate learning phases. This consistency further enhances the visual quality and realism of the HPEs.

The summary of our contributions can be described as follows:
\begin{itemize}
\item We propose the recycle learning strategy for HPE which employs synthetic hand images generated from self-estimated 3D hand information to learn mesh representations. This learning approach facilitates HPE models to enhance generalizability in unseen background and improve the robustness of HPE performance.
\item To maximize the effect of our learning strategy, we introduce a self-correlation loss which encourages the consistency between the outputs from the first training phase and their recycled ones.
\item Our learning approach aims to not only resolve data scarcity but also reduce the discrepancy between quantitative results and their visual qualities.
\item The proposed recycle learning strategy is activated in only training phase. Thus, it achieves improvements in HPE performance without additional computational costs during the inference process.
\end{itemize}
\section{Related Works}
\label{sec:relate}
\subsection{Non-Parametric Methods}
The Non-parametric methods estimate directly hand keypoint positions. However, since a single monocular RGB image has no geometric and depth information, it is challenging to directly employ them for 3D HPE. To overcome this drawback, the conventional non-parametric methods adopt 2D-to-3D lifting paradigm \cite{simon2017hand} or employ depth estimation methods additionally \cite{hand2021survey}. In addition, a multiple stage approach has been introduced \cite{rhd, mueller2018ganerated}, which includes hand segmentation, heat-map based 2D HPE, and 2D-to-3D lifting. However, these 2D-to-3D lifting methods estimate 3D hand keypoint positions from their 2D predictions regardless of detailed spatial information. To alleviate this issue, a latent 2.5D heatmap-based HPE method \cite{Iqbal_2018_ECCV} simultaneously estimates 2D heatmaps for hand keypoint and depth information per pixel in a hand. Thus, 3D hand keypoint positions are estimated by combining both predicted 2D heatmaps and their depth information. However, these non-parametric methods still suffer from the lack of structural representation for hand poses.

\subsection{Parametric Hand Model-Based Methods}
In general, parametric hand model-based methods have been introduced by adopting MANO \cite{mano}, which represents hand poses and meshes using low-dimensional predefined and deformable parameters in the hand structure space. Therefore, the parametric-based method is capable of generating 3D mesh vertices of several hand poses.  METRO \cite{metro} and Mesh Graphormer \cite{mesh_graphormer} are representative methods of the parametric-based 3D HPE using both MANO and transformer \cite{transformer}. Especially, Mesh Graphormer outperforms on HPE adopting similar framework with the GCNN \cite{GCNN}. METRO and Mesh Graphormer employ vertex queries by attaching an image feature to estimate the 3D vertices of a hand mesh. However, when computing the vertex queries, these methods have been designed to burden exceptional computational costs in learning complex 3D hand mesh topology. To overcome this drawback, FastMETRO \cite{fastmetro} utilizes prior knowledge of hand morphological relationships with a disentangled encoder-decoder transformer architecture. In addition, this method employs coarse-to-fine up-sampling strategy to estimate both 3D vertices and keypoints. Its up-sampling strategy adopts a pre-computed matrix utilizing MANO, which exploit the 3D structured representation by considering non-adjacent vertices and keypoints. Therefore, FastMETRO is more faster than previous parametric model-based works while learning intricate the mesh topology more effectively.

\subsection{Datasets for Hand Pose Estimation}
Since datasets have such an enormous impact on the success of 3D hand pose estimation, many researchers spend a lot of effort into acquiring high-quality datasets. A stereo tracking benchmark (STB) \cite{stb} dataset is first and most commonly employed dataset for 3D HPE from a single RGB image. In addition, a panoptic (PAN) dataset is acquired using a dense multi-view capture system consisting of many RGB-D sensors \cite{pan}. Generally, aforementioned hand datasets have been acquired on the real-world. In contrast, a synthetic hand dataset is introduced, called the rendered hand pose dataset (RHD) \cite{rhd}, which renders synthetic hand on random backgrounds. However, RHD contains only hand in a scene without consideration of hand object interaction scenarios. In addition, a HO-3D dataset is acquired incorporating temporal and physical consistencies by utilizing both silhouettes and depth information \cite{ho3d}. Recently, a FreiHAND \cite{freihand} dataset is introduced with 3D annotations in the real-world. This dataset is obtained by iterative semi-automated approach called human-in-the-loop. In FreiHAND, MVNet is applied as a semi-automated hand annotation model, which is trained with bootstrapping procedure. This dataset provides both accurate 3D hand pose and shape annotations since these are acquired employing  multiple RGB cameras. Thus, this dataset is suitable to train the 3D HPE model with considering hand structural information. Moreover, it contains several hand-object interaction scenarios. Consequently, the FreiHAND dataset provides superior generalization of 3D hand capability.

\begin{figure*}
    \centering
    \includegraphics[width=1.0\textwidth]{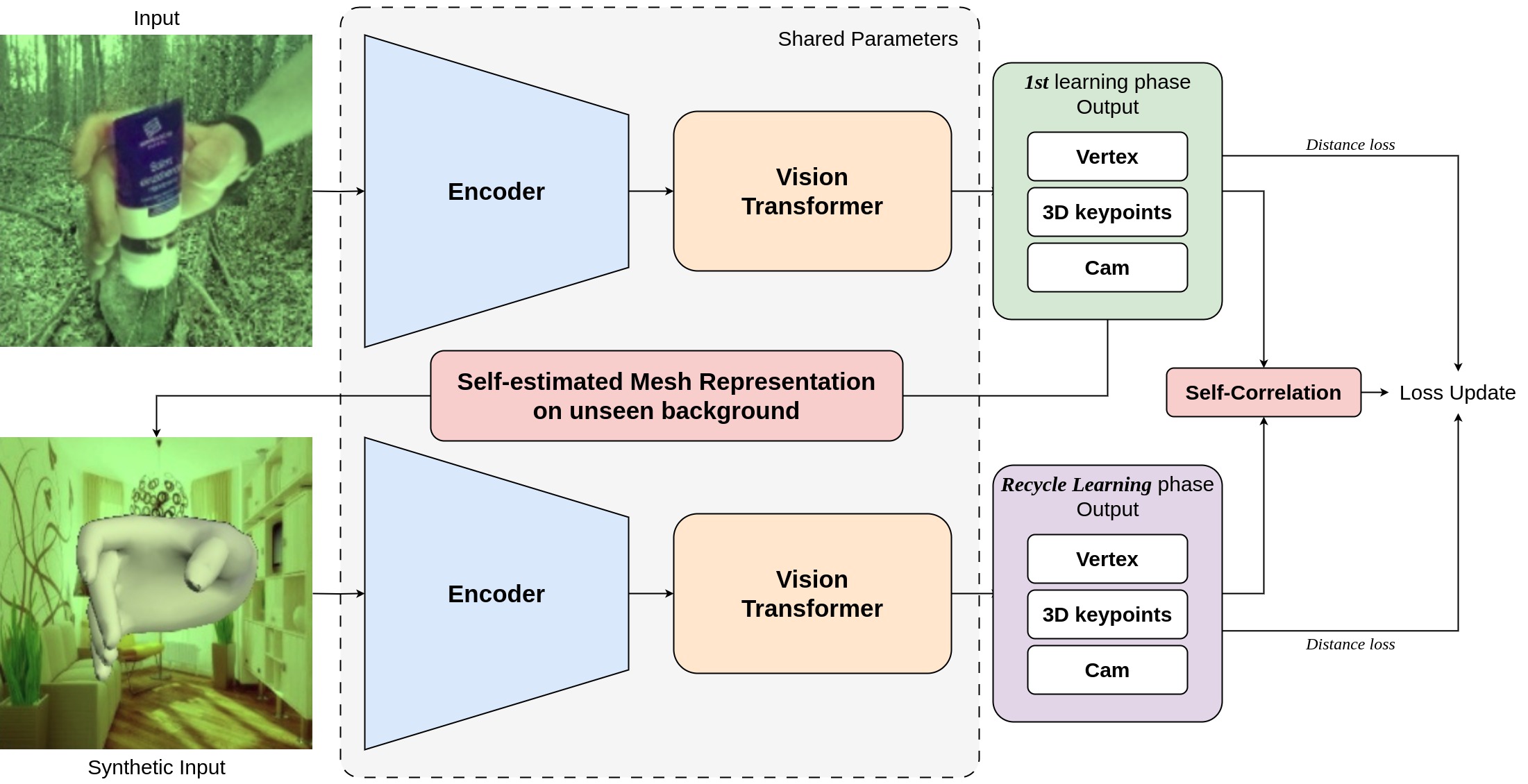}
    \caption{Illustration of the proposed recycle learning strategy. The networks for both original input and synthetic one share their parameters. When rendering a synthetic hand image, LSUN dataset is employed for its background augmentation.}
    \label{architecture}
\end{figure*}
\section{Proposed Method}
As aforementioned in the previous section, the main obstacle of the parametric model-based methods is the model robustness to unseen poses and various backgrounds due to the lack of training data. Thereby, despite achieving the high HPE accuracy, its visual quality is unsatisfactory. To overcome these drawbacks, we propose a novel learning method called recycle learning in this section. We first introduce a recycle learning-based architecture which employs a recursive learning method adopting the model output as input again in the training phase. Thereafter, we describe the self-correlation loss to encourage consistency between the original model output and its recycled one.

\subsection{Recycle Learning}
To introduce the proposed recycle learning approach, we adopt FastMETRO as our baseline which estimates hand mesh representations as not only 3D hand keypoints but also 3D hand vertices. Therefore, we utilize the estimated 3D hand vertices to learn hand mesh representations with the recursive learning method in training phases. To be specific, the recycle learning strategy is comprised of two parts in terms of its training phase; training network with original images and recursive training with rendered synthetic images as illustrated on Figure \ref{architecture}. 

In the first training phase, we obtain 3D hand representations using neural networks from a hand image $I\in \mathbb{R}^{{h}\times {w} \times 3}$ as follows:
\begin{equation}
    \textbf{V}, \textbf{K}_{3D}, \textbf{M}_c = \Phi(I),
\end{equation}
where $\Phi(\cdot)$ is a HPE network, and $\textbf{V} \in \mathbb{R}^{v \times 3}$ and $\textbf{K}_{3D} \in \mathbb{R}^{k \times 3}$ are 3D vertex coordinates and 3D keypoint coordinates separately. In addition, $v$ and $k$ denote the number of vertices and keypoints respectively, and $\textbf{M}_c$ is camera intrinsic parameters. Then, we create 3D hand meshes to represent hand structures and render this synthetic hand into a background image. In this step, the synthetic hand is rendered using an open software called blender \cite{blender}. When generating a synthetic hand image, we employ general backgrounds to consider the robustness of model performance in the real-world scenes. To augment the general backgrounds, we utilize a large-scale scene understanding (LSUN) dataset \cite{lsun} which includes real-world and various scenes. However, the LSUN dataset contains human-existed scenes. Therefore, we filter them out using a YOLO-style detection network \cite{yolov5} to avoid overlapping the synthetic hand and an existing one. After that, these filtered images are randomly selected as backgrounds to synthesize 3D hand vertexes on them. Furthermore, we take average color values of a randomly selected background as the color of synthetic hand to consider the light source and color tones of a rendering image.
Finally, a synthetic hand image $\tilde{I}\in \mathbb{R}^{{h}\times {w} \times 3}$ is generated by utilizing the output of the HPE network.
In the second training phase, we fed the synthetic hand image $\tilde{I}$ into the same HPE network again, then we obtain its 3D hand vertices and 3D hand keypoints with the same manner as follows:
\begin{equation}
    \tilde{\textbf{V}}, \tilde{\textbf{K}}_{3D}, \tilde{\textbf{M}}_{c} = \Phi(\tilde{I}).
\end{equation}
Finally, both outputs estimated from each training phase are compared with their targets to reflect hand mesh representations. We defined this recursive learning strategy as recycle learning.

\subsection{Self-Correlation Loss}
As shown in the Figure \ref{architecture}, we employ the same HPE network $\Phi(\cdot)$ for aforementioned both training phases. From these phases, we acquire a set of 3D hand vertices and 3D hand keypoints separately. Both sets of 3D estimations are compared with the ground-truth of the original hand image to update the HPE network since these are originated from the same hand. However, inconsistent results between them are observed since the HPE network conducts independent estimations from each training phase respectively. To encourage consistency between the original output and its recycled one, the self-correlation loss is applied to both 3D keypoints and 3D vertices respectively as follows:
\begin{equation}
    \mathcal{L}_{corr_k}=\wp(\textbf{K}_{3D},\tilde{\textbf{K}}_{3D}),
    \label{corr_k}
\end{equation}
\begin{equation}
    \mathcal{L}_{corr_v}=\wp(\textbf{V},\tilde{\textbf{V}}),
    \label{corr_v}
\end{equation}
where $\wp(\cdot)$ is a similarity function. In addition, we apply the self-correlation loss to 2D keypoints since camera intrinsic parameters can be estimated by the HPE network or it is provided on ground-truth dataset. To achieve it, 3D keypoints are projected on the image space with $\textbf{M}_c$, and compare 2D keypoints of both outputs similar to Eq. \ref{corr_k} and \ref{corr_v} as follows:
\begin{equation}
    \mathcal{L}_{corr_{proj}}=\wp(\Im(\textbf{K}_{3D}, \textbf{M}_c),\Im(\tilde{\textbf{K}}_{3D}, \tilde{\textbf{M}}_c)), 
    \label{corr_proj}
\end{equation}
where $\Im(\cdot)$ denotes 3D-to-2D keypoint projection on an image using its intrinsic parameters. 
Finally, a total self-correlation loss function can be defined as follows:
\begin{equation}
    \mathcal{L}_{corr}=\mathcal{L}_{corr_k} + \mathcal{L}_{corr_v} + \mathcal{L}_{corr_{proj}}, 
\end{equation}
By adopting this self-correlation manner, we preserve consistency between independent sets of original outputs and their recycled ones.

\subsection{Learning Objectives}
To train the HPE network, we apply additional loss functions to network outputs, and minimize it comparing with their ground-truths. Specifically, 3D coordinates of both keypoints and vertices are updated as follows:
\begin{equation}
    \mathcal{L}_{dist_k} = \left \| (\textbf{K}_{3D},\hat{\textbf{K}}_{3D}) \right \|,
\end{equation}
\begin{equation}
    \mathcal{L}_{dist_v} = \left \| (\textbf{V},\hat{\textbf{V}}) \right \|,
\end{equation}
where $\hat{\textbf{K}}_{3D}$ and $\hat{\textbf{V}}$ are ground-truths of 3D keypoint and 3D vertices coordinates respectively.
In the same manner of Eq. \ref{corr_proj}, 2D keypoint positions is obtained with predictions of 3D keypoint positions and camera intrinsic parameters. Therefore, an additional loss function in the 2D space can be defined as follows:
\begin{equation}
    \mathcal{L}_{dist_{proj}} = \left \| (\Im(\textbf{K}_{3D}, \textbf{M}_{c}),\Im(\hat{\textbf{K}}_{3D}, \hat{\textbf{M}}_{c})) \right \|,
\end{equation}
where $\hat{\textbf{M}}_{c}$ is ground-truth of camera intrinsic parameters in the original image. The total loss function in the first learning phase can be represented as follows:
\begin{equation}
    \mathcal{L}_{ori} = \mathcal{L}_{dist_k} + \mathcal{L}_{dist_v} + \mathcal{L}_{dist_{proj}}.
    \label{loss_ori}
\end{equation}
Its weight balance of each loss function depends on experiment settings of previous works \cite{metro, fastmetro, mesh_graphormer}.
In addition, $\mathcal{L}_{recycle}$, which represents a loss function for predictions of the recycle learning phase, can be defined as the same manner with their original ones since both outputs are originated from the same hand. Finally, the total loss for the proposed recycle learning strategy can be defined as follows:
\begin{equation}
\begin{aligned}
        \mathcal{L}_{total} = \alpha \mathcal{L}_{ori}+ \beta \mathcal{L}_{recycle} + \gamma \mathcal{L}_{corr},
\end{aligned}
\end{equation}
where $\alpha$, $\beta$, and $\gamma$ are weight factors for their corresponding loss functions. When weight factor $\alpha$ is set as one and $\beta$ is set as zero, it is the same learning strategy with an original HPE model. In our learning setting, $\alpha$ and $\beta$ share the same weight value.
\section{Experimental Results}
In this section, we demonstrate the comparisons for our proposed learning method.

\subsection{Dataset}
We adopt FreiHAND dataset \cite{freihand}, which is a challenging and widely adopted dataset for hand pose and mesh estimation. FreiHAND is split into $train$ and $eval$ respectively, and it contains over 134K images. In addition, 21 hand keypoints including wrist and 778 hand vertices are provided, respectively. To evaluate the performance in terms of estimation accuracy, we adopt mean per joint position error using procrustes analysis (PA-MPJPE) \cite{zhou2018monocap} for 3D keypoints and mean per vertex position error (MPVPE) \cite{pavlakos2018learning} for 3D mesh. In addition, F-measure \cite{sasaki2007truth} using two different thresholds (F@5mm and F@15mm) are additionally adopted for analyzing HPE performance. As mentioned in the previous section, we employ human filtered LSUN dataset for background image augmentation in the recycle learning.

\begin{table*}[t]
\centering
\caption{Comparisons of hand pose and mesh estimations with state-of-the-art methods on FreiHAND. $\dagger$ indicates test-time augmentation is applied in the inference phase. In addition, $\downarrow$ indicates that the lower measurement scores are the better performance, $\uparrow$ indicates that the higher measurement scores are the better performance.}
\begin{tabular}{l||c|c||c|c}
\hline
Method           & PA-MPJPE $\downarrow$ & PA-MPVPE $\downarrow$ & F@5mm $\uparrow$ & F@15mm $\uparrow$ \\ \hline \hline
MVNet \cite{freihand}        & -        & 10.7     & 0.529 & 0.935  \\ \hline
Pose2Mesh \cite{pose2mesh}       & 7.7      & 7.8      & 0.674 & 0.969  \\ \hline
I2LMeshNet \cite{moon2020i2l}      & 7.4      & 7.6      & 0.681 & 0.973  \\ \hline
METRO \cite{metro}            & 6.7      & 6.8      & 0.717 & 0.981  \\ \hline
METRO \cite{metro} $\dagger$           & 6.3      & 6.5      & 0.731 & 0.984  \\ \hline
FastMETRO \cite{fastmetro}        & 6.5      & 7.3      & 0.687 & 0.983  \\ \hline
Mesh Graphormer \cite{mesh_graphormer} $\dagger$ & 6.0      & \textbf{5.9}      & \textbf{0.764} & 0.986  \\ \hline \hline
Recycle Learning (Ours) & 6.0      & 6.5      & 0.733 & 0.985  \\ \hline
Recycle Learning (Ours) $\dagger$ & \textbf{5.6}      & 6.2      & 0.752 & \textbf{0.987}  \\ \hline
\end{tabular}
\label{quan_sota}
\end{table*}

\begin{figure*}[!ht]
    \centering
    \includegraphics[width=1.0\textwidth]{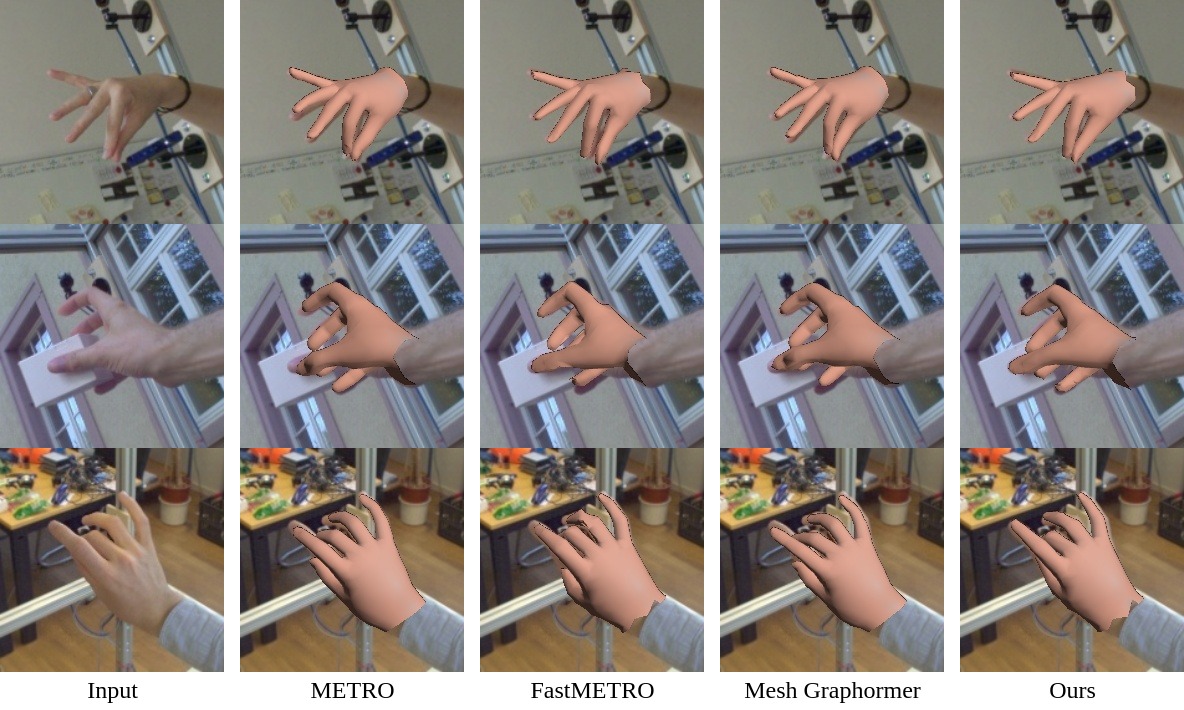}
    \caption{Qualitative comparisons with state-of-the-art methods on FreiHAND. We visualize 3D hand meshes to provide qualitative results. Mesh predictions with our proposed learning method are more fit on challenging hand postures than other ones by recursively learning mesh representations.}
    \label{fig_comp}
\end{figure*}

\subsection{Implementation Details}
To compare HPE performance with conventional methods \cite{freihand, pose2mesh, moon2020i2l, metro, fastmetro, mesh_graphormer}, we adopt FastMETRO as a baseline \cite{fastmetro}. Therefore, coarse-to-fine linear mesh up-sampling is employed with pre-computed matrix obtained from MANO \cite{mano}. To train HPE with the proposed recycle learning strategy on FreiHAND, an initial learning rate is set to 0.0001, and 0.5 is multiplied if PA-MPJPE is unreached to the best performance during 200 epochs. FastMETRO is optimized with adaptive gradient methods such as AdamW \cite{adamw} over 4 NVIDIA V100 GPUs, and the batch sizes is 16 per GPU. On FreiHAND $eval$, our proposed method is performed both subjective and objective quality evaluations. In addition, we provide extensive experiment results by applying our method to various baselines.

\begin{figure*}[t]
    \centering
    \includegraphics[width=1.0\textwidth]{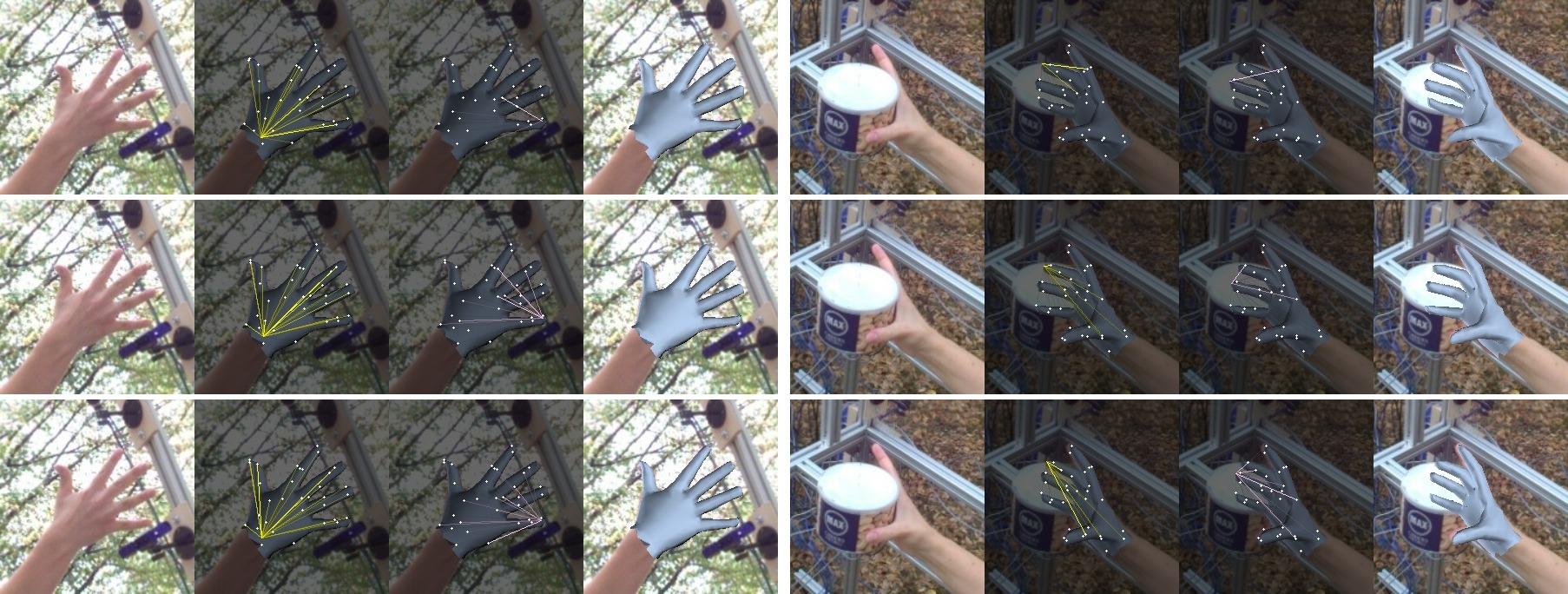}
    \caption{Visualization of attention scores in self-attentions compared with our baseline. First row is results of FastMETRO, and remains are results gradually applying the recycle learning strategy and the self-correlation loss, respectively. Each image contains an original input, two visualized attention scores, and a predicted mesh. Two attention scores in the first column visualize a wrist and pinky finger tip. In the second column, attention scores of a middle and ring finger tip are visualized. The brighter lines represent higher self-attention scores.}
    \label{fig_att}
\end{figure*}

Aforementioned conventional methods employ various similarity measurements in their loss functions such as L1 and L2 norms. Therefore, we adopt a similarity measurement of the self-correlation loss depending on their learning settings. Usually, general loss functions compare the distance between model predictions and their corresponding static ground-truths. Similarly, we adopt L1 and L2 norms as the self-correlation loss, but the distance of keypoint positions between outputs of the first learning and the recycle learning phase. The reason is that a neural network diverges in initial iterations since the distance is large between output from the original image and its self-estimated one. Thus, it is difficult to optimize the self-correlation loss without specific normalization in predictions. To tackle this issue, we normalize both predictions with mean positions of all keypoints. In addition, we align 3D hand meshes form the both predictions by setting their origin as their predicted wrist positions.

\begin{table*}[t]
\centering
\caption{Ablation study of the recycle learning strategy and the self-correlation loss. We gradually apply two proposed methods to various backbones. All networks are trained and evaluated on FreiHAND. In this experiment, test-time augmentation is not applied to these networks.}
\begin{tabular}{l||cc||cc||cc}
\hline
\multicolumn{1}{c||}{\multirow{2}{*}{Backbone}} & \multicolumn{2}{c||}{Original}            & \multicolumn{2}{c||}{+Recycle Learning}   & \multicolumn{2}{c}{+Self-Correlation}   \\ \cline{2-7} 
\multicolumn{1}{c||}{}                          & \multicolumn{1}{c|}{PA-MPJPE} & PA-MPVPE & \multicolumn{1}{c|}{PA-MPJPE} & PA-MPVPE & \multicolumn{1}{c|}{PA-MPJPE} & PA-MPVPE \\ \hline \hline
ResNet101                                       & \multicolumn{1}{c|}{8.53}     & 9.15     & \multicolumn{1}{c|}{8.35}     & 8.96     & \multicolumn{1}{c|}{8.32}     & 8.93     \\ \hline
HRNet-W64                                       & \multicolumn{1}{c|}{7.04}     & 7.80     & \multicolumn{1}{c|}{6.40}     & 7.18     & \multicolumn{1}{c|}{6.39}     & 7.17     \\ \hline
FastMETRO                                       & \multicolumn{1}{c|}{6.56}     & 7.34     & \multicolumn{1}{c|}{6.18}     & 6.72     & \multicolumn{1}{c|}{6.02}     & 6.53     \\ \hline
\end{tabular}
\label{ablation}
\end{table*}

\begin{table}[t]
\centering
\caption{Verification of rendering quality. MVNet and FastMETRO are trained and evaluated on a real-world dataset. In contrast, Synthetic Image is trained on hand-rendered images and evaluated on the real-world dataset.}
\begin{tabular}{l||c}
\hline
\multicolumn{1}{c||}{Method} & PA-MPVPE \\ \hline \hline
MVNet                     & 10.7     \\ \hline
FastMETRO                    & 7.3      \\ \hline
Synthetic Image              & 8.8      \\ \hline
\end{tabular}
\label{rendering_quality}
\end{table}

\subsection{State-of-the-art Analysis}
We compare our method with state-of-the-art HPE methods on the FreiHAND dataset. Table \ref{quan_sota} describes quantitative comparisons in terms of hand pose and mesh estimation. In these experiments, test-time augmentation (TTA) is applied to METRO, Mesh Graphormer, and our method. As described in Table \ref{quan_sota}, it is obvious that our method shows comparable performance with current state-of-the-art methods. Specifically, our method achieves the best performance on PA-MPJPE with 5.6 mm while showing an acceptable result on PA-MPVPE. In addition, our method without TTA still maintains the state-of-the-art performance on 3D keypoint estimations, even our adopting baseline demands less parameters than Mesh Graphormer \cite{fastmetro}. Furthermore, Figure \ref{fig_comp} provides qualitative comparisons with previous methods by rendering predicted meshes. As shown in this figure, our method generates higher-quality hand meshes compared with state-of-the-art methods. Specifically, it is obvious from this figure that our proposed learning method is more robust to unusual hand poses and occluded hands compared with other methods. The reason is that the recycle learning strategy provides non-occluded synthetic hand images utilizing self-estimated hand representation in the recursive learning phase. In addition, the self-correlation loss minimizes the distance of mesh predictions between a occluded hand in the original image and its non-occluded hand in a synthetic one. Thus, our method shows superior performance on challenging hand poses.

\begin{figure*}[t]
    \centering
    \includegraphics[width=1.0\textwidth]{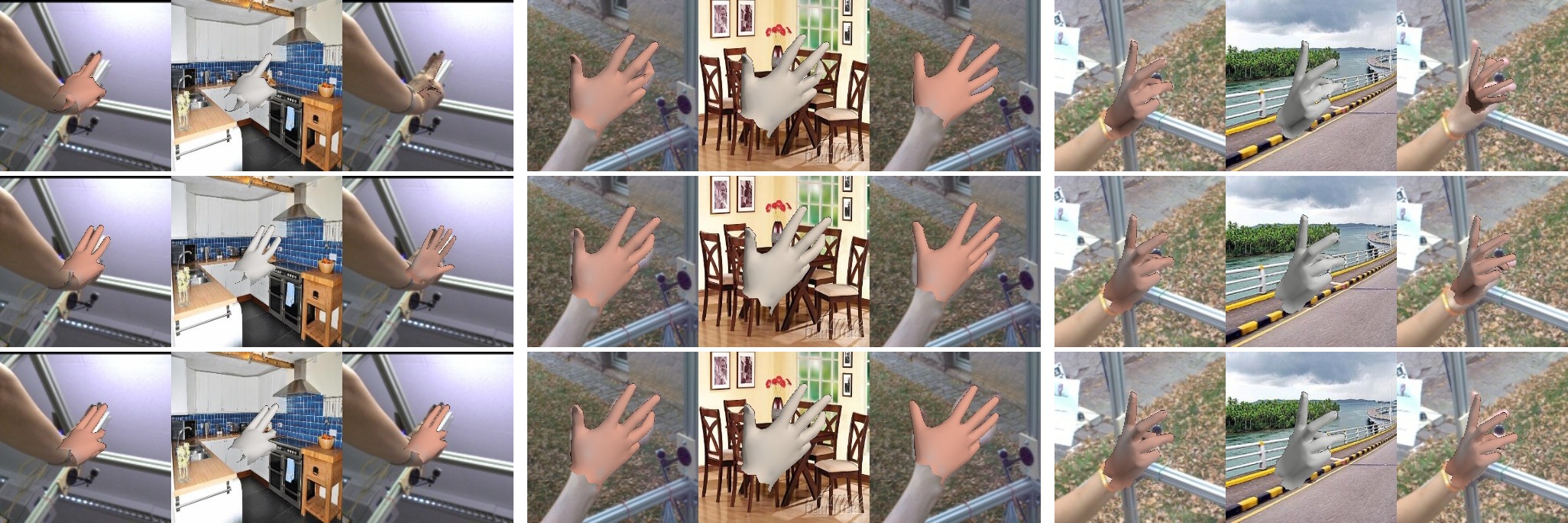}
    \caption{Visualization of hand mesh predictions compared with our baseline. First row is results of FastMETRO, and remains are results gradually applying the recycle learning strategy and the self-correlation loss, respectively. Each image contains an output mesh from an original input, a rendering hand mesh of the original output on unseen background, and a predicted hand mesh from a rendered image.}
    \label{fig_corr}
\end{figure*}

\subsection{Analysis of Subjective and Objective Quality}
\textbf{Rendering Quality}
To verify hand rendering quality in the recycle learning strategy, we train a HPE network on only synthetic hand images and evaluate it on a real-world dataset. To conduct this experiment, we render hand meshes with their ground-truths. After that, the HPE network is updated with only the recycle learning loss $\mathcal{L}_{recycle}$. In addition, we adopt FastMETRO as a baseline network and FreiHAND as a real-world evaluation dataset. As described in Table \ref{rendering_quality}, Synthetic Image shows comparative performance on PA-MPVPE. Concretely, Synthetic Image shows only 1.5 mm difference in PA-MPVPE compared with FastMETRO trained on the real-world dataset although these two networks share the similar architecture and learning settings. Furthermore, this network outperforms MVNet \cite{freihand}, which is employed to acquire the FreiHAND dataset, although MVNet utilizes multi-view images capturing a single hand on the real-world scene to fit its deformable hand shape model. These results demonstrate that a quality of synthetic hand images are reasonable to train a hand pose and mesh estimation network for real-world. Thus, the recycle learning strategy ensures improvement of model performance on hand pose and mesh estimations.

\textbf{Adaptation Capacity}
To demonstrate our proposed learning method, we train and evaluate various networks by gradually applying the recycle learning strategy and the self-correlation loss. We adopt ResNet101 \cite{resnet}, HRNet-W64 \cite{hrnet}, and FastMETRO as backbone networks, which are widespread and state-of-the-art methods in vision and pose estimation tasks. To implement this experiment, we change the final layer of ResNet101 and HRNet-W64 from classification to hand pose and mesh estimations. In other networks, we follow their original architecture settings. In addition, we set implementation setting of ResNet101 and HRNet-W64 similar to FastMETRO. As shown in Table \ref{ablation}, performances of PA-MPJPE and PA-MPVPE in all networks are improved by employing our proposed learning method. Concretely, the recycle learning strategy is effective both in convolution-based networks (i.e., ResNet101 and HRNet-W64) and transformer-based one (i.e., FastMETRO). Similarly, the self-correlation loss enhances performance of these networks on hand pose and mesh estimations. These results indicate that our learning method facilitates HPE networks to effectively learn hand mesh representations through recursive learning using their self-estimated results.

\textbf{Mesh Representation and Consistency}
To verify the effectiveness of our learning strategy, we visualize attention scores in self-attentions between a specific hand keypoint and non-adjacent mesh vertices. We calculate visualized scores by averaging attention scores from attention heads of multi-head self-attention modules in transformer blocks. For this experiment, we gradually apply the recycle learning strategy and the self-correlation loss to our baseline. As shown in Figure \ref{fig_att}, our proposed method promotes the transformer blocks to capture non-local relationships between specific hand keypoint and vertices. Specifically, attention scores of a pinky finger tip in the first column are more highlighted, which enhances mesh predictions in the pinky finger tip. The reason is that the recycle learning strategy conducts self-enhancement by recursively learning additional mesh representations from own output. Furthermore, non-local relationships between occluded keypoints and vertices are improved in the second column of Figure \ref{fig_att} since a non-occluded synthetic hand is provided to interpret its mesh representations. Consequently, our learning strategy enhances the robustness of hand pose and mesh estimation in challenging hand scenes.

In addition, we visualize hand meshes of output from the original image and its synthetic hand on an unseen background to demonstrate visual effects of our proposed method. As shown in Figure \ref{fig_corr}, our learning method visually improves predictions of hand meshes compared with our baseline. In the first column, hand mesh prediction is challenging since a light source is irregularly highlighted and a hand is overlapped with objects. Thus, our baseline yields an inaccurate result on the hand mesh prediction. In contrast, our learning method gradually improves the hand mesh prediction. In addtion, similar tendencies are observed in vertices of a ring finger in images of the second and third columns. Furthermore, the self-correlation loss encourages consistency between output vertices from the original output and its rendered one as shown in the second and third rows of Figure \ref{fig_corr}. Thus, it is demonstrated that the self-correlation loss guides the hand pose and mesh estimation model to extract the same hand representations between an original hand and its synthetic hand.
\section{Conclusion}
\label{sec:conclusion}
In this paper, we propose the mesh represented recycle learning strategy for parametric model-based 3D hand pose and mesh estimation. To be specific, a hand and mesh estimation model first predicts 3D hand keypoints and vertices with real-world hand images. After that, synthetic hand images are generated with self-estimated hand mesh representations. Finally, these images are fed into the same model again to learn its self-estimated mesh representations. Therefore, the proposed learning strategy achieves performance improvement for both the quantitative results and visual qualities by reinforcing synthetic mesh representation. To promote efficiency of the recycle learning strategy, we further propose the self-correlation loss which encourages consistency between original model output and its recycled one. Thus, the self-correlation loss maximizes both the accuracy and reliability of our learning strategy. Consequently, our proposed method facilities a learning model to perform self-refinement on hand pose and mesh estimation by effectively learning mesh representation from its own output. Moreover, our learning strategy achieves performance improvement on hand pose and mesh estimation without any extra execution time during the inference.

{
    \small
    \bibliographystyle{ieeenat_fullname}
    \bibliography{main}

\begin{thebibliography}{34}
\providecommand{\natexlab}[1]{#1}
\providecommand{\url}[1]{\texttt{#1}}
\expandafter\ifx\csname urlstyle\endcsname\relax
  \providecommand{\doi}[1]{doi: #1}\else
  \providecommand{\doi}{doi: \begingroup \urlstyle{rm}\Url}\fi

\bibitem[Baek et~al.(2019)Baek, Kim, and Kim]{Baek_2019_CVPR}
Seungryul Baek, Kwang~In Kim, and Tae-Kyun Kim.
\newblock Pushing the envelope for rgb-based dense 3d hand pose estimation via neural rendering.
\newblock In \emph{Proceedings of the IEEE/CVF Conference on Computer Vision and Pattern Recognition (CVPR)}, 2019.

\bibitem[Cho et~al.(2022)Cho, Youwang, and Oh]{fastmetro}
Junhyeong Cho, Kim Youwang, and Tae-Hyun Oh.
\newblock Cross-attention of disentangled modalities for 3d human mesh recovery with transformers.
\newblock In \emph{European Conference on Computer Vision}, pages 342--359. Springer, 2022.

\bibitem[Choi et~al.(2020)Choi, Moon, and Lee]{pose2mesh}
Hongsuk Choi, Gyeongsik Moon, and Kyoung~Mu Lee.
\newblock Pose2mesh: Graph convolutional network for 3d human pose and mesh recovery from a 2d human pose.
\newblock In \emph{Computer Vision--ECCV 2020: 16th European Conference, Glasgow, UK, August 23--28, 2020, Proceedings, Part VII 16}, pages 769--787. Springer, 2020.

\bibitem[Community(2018)]{blender}
Blender~Online Community.
\newblock \emph{Blender - a 3D modelling and rendering package}.
\newblock Blender Foundation, Stichting Blender Foundation, Amsterdam, 2018.

\bibitem[Doosti(2019)]{hand2019survey}
Bardia Doosti.
\newblock Hand pose estimation: A survey.
\newblock \emph{arXiv preprint arXiv:1903.01013}, 2019.

\bibitem[Gomez-Donoso et~al.(2019)Gomez-Donoso, Orts-Escolano, and Cazorla]{gomez2019large}
Francisco Gomez-Donoso, Sergio Orts-Escolano, and Miguel Cazorla.
\newblock Large-scale multiview 3d hand pose dataset.
\newblock \emph{Image and Vision Computing}, 81:\penalty0 25--33, 2019.

\bibitem[Grady et~al.(2021)Grady, Tang, Twigg, Vo, Brahmbhatt, and Kemp]{grady2021contactopt}
Patrick Grady, Chengcheng Tang, Christopher~D Twigg, Minh Vo, Samarth Brahmbhatt, and Charles~C Kemp.
\newblock Contactopt: Optimizing contact to improve grasps.
\newblock In \emph{Proceedings of the IEEE/CVF Conference on Computer Vision and Pattern Recognition}, pages 1471--1481, 2021.

\bibitem[Hampali et~al.(2020)Hampali, Rad, Oberweger, and Lepetit]{ho3d}
Shreyas Hampali, Mahdi Rad, Markus Oberweger, and Vincent Lepetit.
\newblock Honnotate: A method for 3d annotation of hand and object poses.
\newblock In \emph{Proceedings of the IEEE/CVF conference on computer vision and pattern recognition}, pages 3196--3206, 2020.

\bibitem[He et~al.(2016)He, Zhang, Ren, and Sun]{resnet}
Kaiming He, Xiangyu Zhang, Shaoqing Ren, and Jian Sun.
\newblock Deep residual learning for image recognition.
\newblock In \emph{Proceedings of the IEEE conference on computer vision and pattern recognition}, pages 770--778, 2016.

\bibitem[Huang et~al.(2021)Huang, Zhang, Guo, Xiao, Cao, and Yuan]{hand2021survey}
Lin Huang, Boshen Zhang, Zhilin Guo, Yang Xiao, Zhiguo Cao, and Junsong Yuan.
\newblock Survey on depth and rgb image-based 3d hand shape and pose estimation.
\newblock \emph{Virtual Reality \& Intelligent Hardware}, 3\penalty0 (3):\penalty0 207--234, 2021.

\bibitem[Iqbal et~al.(2018)Iqbal, Molchanov, Gall, and Kautz]{Iqbal_2018_ECCV}
Umar Iqbal, Pavlo Molchanov, Thomas Breuel~Juergen Gall, and Jan Kautz.
\newblock Hand pose estimation via latent 2.5d heatmap regression.
\newblock In \emph{Proceedings of the European Conference on Computer Vision (ECCV)}, 2018.

\bibitem[Jocher et~al.(2020)Jocher, Changyu, Hogan, Yu, Rai, Sullivan, et~al.]{yolov5}
Glenn Jocher, Liu Changyu, Adam Hogan, Lijun Yu, Prashant Rai, Trevor Sullivan, et~al.
\newblock ultralytics/yolov5: Initial release.
\newblock \emph{Zenodo}, 2020.

\bibitem[Joo et~al.(2018)Joo, Simon, and Sheikh]{pan}
Hanbyul Joo, Tomas Simon, and Yaser Sheikh.
\newblock Total capture: A 3d deformation model for tracking faces, hands, and bodies.
\newblock In \emph{Proceedings of the IEEE conference on computer vision and pattern recognition}, pages 8320--8329, 2018.

\bibitem[Kolotouros et~al.(2019)Kolotouros, Pavlakos, and Daniilidis]{GCNN}
Nikos Kolotouros, Georgios Pavlakos, and Kostas Daniilidis.
\newblock Convolutional mesh regression for single-image human shape reconstruction.
\newblock In \emph{Proceedings of the IEEE/CVF Conference on Computer Vision and Pattern Recognition}, pages 4501--4510, 2019.

\bibitem[Lin et~al.(2021{\natexlab{a}})Lin, Wang, and Liu]{mesh_graphormer}
Kevin Lin, Lijuan Wang, and Zicheng Liu.
\newblock Mesh graphormer.
\newblock In \emph{Proceedings of the IEEE/CVF international conference on computer vision}, pages 12939--12948, 2021{\natexlab{a}}.

\bibitem[Lin et~al.(2021{\natexlab{b}})Lin, Wang, and Liu]{metro}
Kevin Lin, Lijuan Wang, and Zicheng Liu.
\newblock End-to-end human pose and mesh reconstruction with transformers.
\newblock In \emph{Proceedings of the IEEE/CVF conference on computer vision and pattern recognition}, pages 1954--1963, 2021{\natexlab{b}}.

\bibitem[Liu et~al.(2021)Liu, Jiang, Xu, Liu, and Wang]{liu2021semi}
Shaowei Liu, Hanwen Jiang, Jiarui Xu, Sifei Liu, and Xiaolong Wang.
\newblock Semi-supervised 3d hand-object poses estimation with interactions in time.
\newblock In \emph{Proceedings of the IEEE/CVF Conference on Computer Vision and Pattern Recognition}, pages 14687--14697, 2021.

\bibitem[Loshchilov and Hutter(2017)]{adamw}
Ilya Loshchilov and Frank Hutter.
\newblock Decoupled weight decay regularization.
\newblock \emph{arXiv preprint arXiv:1711.05101}, 2017.

\bibitem[Moon and Lee(2020)]{moon2020i2l}
Gyeongsik Moon and Kyoung~Mu Lee.
\newblock I2l-meshnet: Image-to-lixel prediction network for accurate 3d human pose and mesh estimation from a single rgb image.
\newblock In \emph{Computer Vision--ECCV 2020: 16th European Conference, Glasgow, UK, August 23--28, 2020, Proceedings, Part VII 16}, pages 752--768. Springer, 2020.

\bibitem[Mueller et~al.(2018)Mueller, Bernard, Sotnychenko, Mehta, Sridhar, Casas, and Theobalt]{mueller2018ganerated}
Franziska Mueller, Florian Bernard, Oleksandr Sotnychenko, Dushyant Mehta, Srinath Sridhar, Dan Casas, and Christian Theobalt.
\newblock Ganerated hands for real-time 3d hand tracking from monocular rgb.
\newblock In \emph{Proceedings of the IEEE conference on computer vision and pattern recognition}, pages 49--59, 2018.

\bibitem[Ohkawa et~al.(2023)Ohkawa, Furuta, and Sato]{ohkawa2023efficient}
Takehiko Ohkawa, Ryosuke Furuta, and Yoichi Sato.
\newblock Efficient annotation and learning for 3d hand pose estimation: a survey.
\newblock \emph{International Journal of Computer Vision}, pages 1--14, 2023.

\bibitem[Pavlakos et~al.(2018)Pavlakos, Zhu, Zhou, and Daniilidis]{pavlakos2018learning}
Georgios Pavlakos, Luyang Zhu, Xiaowei Zhou, and Kostas Daniilidis.
\newblock Learning to estimate 3d human pose and shape from a single color image.
\newblock In \emph{Proceedings of the IEEE conference on computer vision and pattern recognition}, pages 459--468, 2018.

\bibitem[Romero et~al.(2022)Romero, Tzionas, and Black]{mano}
Javier Romero, Dimitrios Tzionas, and Michael~J Black.
\newblock Embodied hands: Modeling and capturing hands and bodies together.
\newblock \emph{arXiv preprint arXiv:2201.02610}, 2022.

\bibitem[Sasaki et~al.(2007)]{sasaki2007truth}
Yutaka Sasaki et~al.
\newblock The truth of the f-measure.
\newblock \emph{Teach tutor mater}, 1\penalty0 (5):\penalty0 1--5, 2007.

\bibitem[Simon et~al.(2017)Simon, Joo, Matthews, and Sheikh]{simon2017hand}
Tomas Simon, Hanbyul Joo, Iain Matthews, and Yaser Sheikh.
\newblock Hand keypoint detection in single images using multiview bootstrapping.
\newblock In \emph{Proceedings of the IEEE conference on Computer Vision and Pattern Recognition}, pages 1145--1153, 2017.

\bibitem[Vaswani et~al.(2017)Vaswani, Shazeer, Parmar, Uszkoreit, Jones, Gomez, Kaiser, and Polosukhin]{transformer}
Ashish Vaswani, Noam Shazeer, Niki Parmar, Jakob Uszkoreit, Llion Jones, Aidan~N Gomez, {\L}ukasz Kaiser, and Illia Polosukhin.
\newblock Attention is all you need.
\newblock \emph{Advances in neural information processing systems}, 30, 2017.

\bibitem[Wan et~al.(2018)Wan, Probst, Van~Gool, and Yao]{wan2018dense}
Chengde Wan, Thomas Probst, Luc Van~Gool, and Angela Yao.
\newblock Dense 3d regression for hand pose estimation.
\newblock In \emph{Proceedings of the IEEE conference on computer vision and pattern recognition}, pages 5147--5156, 2018.

\bibitem[Wang et~al.(2020)Wang, Sun, Cheng, Jiang, Deng, Zhao, Liu, Mu, Tan, Wang, et~al.]{hrnet}
Jingdong Wang, Ke Sun, Tianheng Cheng, Borui Jiang, Chaorui Deng, Yang Zhao, Dong Liu, Yadong Mu, Mingkui Tan, Xinggang Wang, et~al.
\newblock Deep high-resolution representation learning for visual recognition.
\newblock \emph{IEEE transactions on pattern analysis and machine intelligence}, 43\penalty0 (10):\penalty0 3349--3364, 2020.

\bibitem[Ye and Kim(2018)]{ye2018occlusion}
Qi Ye and Tae-Kyun Kim.
\newblock Occlusion-aware hand pose estimation using hierarchical mixture density network.
\newblock In \emph{Proceedings of the European conference on computer vision (ECCV)}, pages 801--817, 2018.

\bibitem[Yu et~al.(2015)Yu, Seff, Zhang, Song, Funkhouser, and Xiao]{lsun}
Fisher Yu, Ari Seff, Yinda Zhang, Shuran Song, Thomas Funkhouser, and Jianxiong Xiao.
\newblock Lsun: Construction of a large-scale image dataset using deep learning with humans in the loop.
\newblock \emph{arXiv preprint arXiv:1506.03365}, 2015.

\bibitem[Zhang et~al.(2016)Zhang, Jiao, Chen, Qu, Xu, and Yang]{stb}
Jiawei Zhang, Jianbo Jiao, Mingliang Chen, Liangqiong Qu, Xiaobin Xu, and Qingxiong Yang.
\newblock 3d hand pose tracking and estimation using stereo matching.
\newblock \emph{arXiv preprint arXiv:1610.07214}, 2016.

\bibitem[Zhou et~al.(2018)Zhou, Zhu, Pavlakos, Leonardos, Derpanis, and Daniilidis]{zhou2018monocap}
Xiaowei Zhou, Menglong Zhu, Georgios Pavlakos, Spyridon Leonardos, Konstantinos~G Derpanis, and Kostas Daniilidis.
\newblock Monocap: Monocular human motion capture using a cnn coupled with a geometric prior.
\newblock \emph{IEEE transactions on pattern analysis and machine intelligence}, 41\penalty0 (4):\penalty0 901--914, 2018.

\bibitem[Zimmermann and Brox(2017)]{rhd}
Christian Zimmermann and Thomas Brox.
\newblock Learning to estimate 3d hand pose from single rgb images.
\newblock In \emph{Proceedings of the IEEE international conference on computer vision}, pages 4903--4911, 2017.

\bibitem[Zimmermann et~al.(2019)Zimmermann, Ceylan, Yang, Russell, Argus, and Brox]{freihand}
Christian Zimmermann, Duygu Ceylan, Jimei Yang, Bryan Russell, Max Argus, and Thomas Brox.
\newblock Freihand: A dataset for markerless capture of hand pose and shape from single rgb images.
\newblock In \emph{Proceedings of the IEEE/CVF International Conference on Computer Vision}, pages 813--822, 2019.

\end{thebibliography}
}


\end{document}